\documentclass{ecai} 

\usepackage{latexsym}
\usepackage{amssymb}
\usepackage{amsmath}
\usepackage{amsthm}
\usepackage{booktabs}
\usepackage{enumitem}
\usepackage{graphicx}
\graphicspath{{}}
\usepackage{color}
\usepackage{multicol}
\usepackage{multirow}
\usepackage{float}

\newtheorem{theorem}{Theorem}
\newtheorem{lemma}[theorem]{Lemma}

\newcommand{\BibTeX}{B\kern-.05em{\sc i\kern-.025em b}\kern-.08em\TeX}

\begin{document}


\begin{frontmatter}


\paperid{2148} 


\title{Reducing Oversmoothing through Informed Weight Initialization in Graph Neural Networks}


\author[A,B]{\fnms{Dimitrios}~\snm{Kelesis}\thanks{Corresponding Author. Email: dkelesis@iit.demokritos.gr}}
\author[B]{\fnms{Dimitris}~\snm{Fotakis}}
\author[A]{\fnms{Georgios}~\snm{Paliouras}} 

\address[A]{School of Electrical Engineering and Computer Science, National Technical University of Athens, Greece}
\address[B]{National Center for Scientific Research ``Demokritos'', Greece}


\begin{abstract}
In this work, we generalize the ideas of Kaiming initialization to Graph Neural Networks (GNNs) and propose a new scheme (G-Init) that reduces oversmoothing, leading to very good results in node and graph classification tasks. GNNs are commonly initialized using methods designed for other types of Neural Networks, overlooking the underlying graph topology. We analyze theoretically the variance of signals flowing forward and gradients flowing backward in the class of convolutional GNNs. We then simplify our analysis to the case of the GCN and propose a new initialization method. Our results indicate that the new method (G-Init) reduces oversmoothing in deep GNNs, facilitating their effective use. Experimental validation supports our theoretical findings, demonstrating the advantages of deep networks in scenarios with no feature information for unlabeled nodes (i.e., ``cold start'' scenario).
\end{abstract}

\end{frontmatter}


\section{Introduction}
Weight initialization has been shown to play an important role in the training of neural networks \citep{He, Xavier}. The choice of initial weight values, when made in an informed manner, can significantly impact the training convergence and the final performance of the model. Informed weight initialization methods aim to balance between avoiding convergence to suboptimal solutions and preventing exploding or vanishing gradients. The most prominent techniques take into account the network's architecture and the activation functions, aiming to stabilize the variance of signals flowing forward and gradients flowing backward through the model.\\
Despite their success in Feed Forward Networks (FFNs), the aforementioned weight initialization methods are not directly applicable to GNNs. The underlying graph structure and message passing in GNNs affect the information flow between network neurons, which in turn impacts the variance of the flowing signals and gradients.\\
Meanwhile, the interest in deep GNNs has been increasing \citep{GCNII, Dropedge}, but common GNN models, like the Graph Convolutional Network (GCN) quickly lose their ability to create informative node representations as their depth increases \citep{Li_et}. This is due to the neighborhood aggregation occurring in every layer of the model, which is a type of Laplacian smoothing. Stacking multiple layers in GNNs leads to the oversmoothing phenomenon \citep{Suzuki, A_note}, where node representations are indistinguishable, which in turn deteriorates the model's performance.\\
Several methods have been proposed to alleviate oversmoothing and enable deep GNNs \citep{dirichlet_energy, Pairnorm}, but none of them considers the effect of weight initialization. The aim of this paper is to propose a weight initialization that is suitable for GNNs and investigate the effect weight initialization can have to oversmoothing. In particular, we generalize the analysis of \citet{He} to GNNs and present results about the variance of signals and gradients flowing inside the network. We analyze the case of a general GNN model, which may combine graph convolution, residual connections and skip connections, aiming to cover the majority of existing graph convolutional models. Leveraging these theoretical results, we then focus on the simpler case of a GCN and we propose a novel weight initialization method (G-Init), that stabilizes variance and reduces oversmoothing. We attribute this effect to the initial largest singular values of the weight matrices, which may influence the largest singular values after convergence. These, in turn, are known to be related to oversmoothing \citep{Suzuki}. Finally, experiments on a diverse range of datasets verify our theoretical results.\\
Hence, the main contributions of this paper are as follows:\\

\noindent \textbullet \hspace{.01cm} \textbf{Theoretical analysis of weight initialization for GNNs:} We generalize the analysis of \citet{He} to convolutional GNNs. We investigate a parametric GNN encompassing variations with and without residual connections, skip connections, identity addition to the weight matrix and controlled effects of graph convolution and weight matrix. We derive the respective formulas about the variance of the forward signals and the backward gradients within the model.\\

\noindent \textbullet \hspace{.01cm} \textbf{A new weight initialization method (G-Init):} We propose a new way to initialize GNN weight matrices, focusing on the initialization of GCN. Our method aims to stabilize the variance and capitalize on its relationship with oversmoothing reduction.\\

\noindent \textbullet \hspace{.01cm} \textbf{Experimental confirmation of the benefits of G-Init:} We conduct experiments utilizing deep GCNs with up to 64 layers across 8 datasets for node classification tasks and show that the proposed initialization reduces oversmoothing. Additionally, we demonstrate the benefits of G-Init, in the presence of the ``cold start'' situation, where node features are available only for the labeled nodes. Additionally, we extend our experiments to graph classification tasks, in which G-Init outperforms standard initialization methods.

\section{Notations and Preliminaries}
\subsection{Notations}
We will focus on the common task of semi-supervised node classification on a graph. The graph under investigation is G($\mathbb{V,E}$,X), with $|\mathbb{V}| = N$ nodes $u_i \in \mathbb{V}$, edges $(u_i, u_j) \in \mathbb{E}$ and $X=[x_1,...,x_N]^T \in \mathbb{R}^{N \times C}$ the initial node features. The edges form an adjacency matrix $A \in \mathbb{R}^{N \times N}$ where edge $(u_i, u_j)$ is associated with element $A_{i,j}$. $A_{i,j}$ can take arbitrary real values indicating the weight (strength) of edge $(u_i, u_j)$. Node degrees are represented through a diagonal matrix $D \in \mathbb{R}^{N \times N}$, where each element $d_i$ represents the sum of edge weights connected to node i. During training, only the labels of a subset $V_l \in \mathbb{V}$ are available. The task is to learn a node classifier, that predicts the label of each node using the graph topology and the given feature vectors.\\
\textbf{GCN} originally proposed by \cite{Kipf_gcn}, utilizes a feed forward propagation as:
\begin{equation}\label{eq:gcn}
    H^{(l+1)} = \sigma (\hat{A}H^{(l)}W^{(l)}),
\end{equation}
where $H^{(l)} = [h^{(l)}_1,...,h^{(l)}_N]$ are node representations (or hidden vectors or embeddings) of the $l$-th layer, with $h^{(l)}_i$ standing for the hidden representation of node i; $\hat{A} = \hat{D}^{-1/2} (A+I)\hat{D}^{-1/2}$ denotes the augmented symmetrically normalized adjacency matrix after self-loop addition, where $\hat{D}$ corresponds to the degree matrix; $\sigma(\cdot)$ is a nonlinear element-wise function, i.e. the activation function, which is typically ReLU; and $W^{(l)}$ is the trainable weight matrix of the $l$-th layer. \\
Extending the GCN formula to a parametric and more generic GNN we arrive at:
\begin{align*}
    H^{(l+1)} = \sigma\big(\big(\alpha \hat{A}H^{(l)} + \beta H^{(0)} + \gamma H^{(l-1)} \big) \times
\end{align*}
\begin{equation}\label{eq:general_gnn}
    \big(\delta W^{(l)} + \epsilon I \big) \big),
\end{equation}
where $\alpha, \beta, \gamma, \delta \text{ and } \epsilon$ are preselected parameters that determine the convolutional GNN architecture (e.g., if $\alpha=\delta=1, \beta=\gamma=\epsilon=0$ we have a GCN, while if $\alpha=0.9, \beta=0.1, \gamma=0, \delta = \lambda / l, \epsilon = 1 - \delta$  we arrive at GCNII \citep{GCNII}). We observe that, $\alpha$ and $\delta$ control the importance of graph convolution and the effect of the weight matrix respectively. Furthermore, $\beta, \gamma$ and $\epsilon$ dictate the existence or absence of residual connections, skip connections and identity addition respectively.\\
GNNs are also commonly used for graph classification tasks. Such tasks involve multiple graphs of arbitrary sizes, each associated with a label that needs to be predicted. To achieve this, GNNs are slightly extended with the addition of a $READOUT$ function. That function takes as input the final node embeddings and aggregates them to produce a graph representation. It is common practice to use average pooling as a $READOUT$ function. We will follow the same approach in our experiments throughout this work. 

\subsection{Understanding Oversmoothing}
\citet{Li_et} showed that graph convolution is a type of Laplacian smoothing, caused by repeated neighborhood aggregation. GNNs take advantage of the smoothing process to create similar node representations within each (graph) cluster, i.e. densely connected group of nodes, which in turn improves the performance on semi-supervised tasks on graphs. Increasing the depth of the model leads to more extensive smoothing and ultimately oversmoothing of node representations, i.e. node representations become too similar and much of the initial information is lost.\\
\citet{Suzuki} have generalized the idea in \citet{Li_et} considering also, that the ReLU activation function maps to a positive cone. They explain oversmoothing as a convergence to a subspace, and provide an estimate of the speed of convergence to this subspace. That speed is expressed as the distance of node representations from the oversmoothing subspace $M$ (details can be found in \citet{Suzuki}).
\begin{theorem}[\citet{Suzuki}]\label{theo:suzuki}
Let $s_l=\prod\limits_{h=1}^{H_l}{s_{lh}}$ where $s_{lh}$ is the largest singular value of weight matrix $W_{lh}$ and s = $sup_{l\in \mathbb{N}_+} s_{l}$. Then $d_M(X^{(l)}) = O((s\lambda)^l)$, where l is the layer number and if $s\lambda < 1$ the distance from the oversmoothing subspace exponentially approaches
zero. Where $\lambda$ is the smallest non-zero eigenvalue of $I - \hat{A}$.
\end{theorem}
\noindent Based on Theorem \ref{theo:suzuki}, deep GCNs are expected to suffer from oversmoothing. Since the topology is predefined, to reduce oversmoothing the model would have to maintain a high value of the product of the largest singular values of the weight matrices.

\subsection{Weight Initialization}
Before training, all entries of the weight matrices are sampled from a probability distribution. Selecting the appropriate distribution is very important. The most commonly used initialization methods (i.e., \citet{Xavier} and \citet{He}) focus on the stabilization of the variance across layers. They aim to stabilize both the variance of the signals flowing forward and of the gradients which flow backwards. The aforementioned methods either use uniform or zero-mean Gaussian distributions. When Gaussian distributions are used, their variance plays a crucial role. In particular, \citet{Xavier} proposed to initialize the weights, using a zero-mean Gaussian with variance equal to $1/ n_l$, where $n_l$ is the dimensionality of the $l$-th layer. Moving one step further \citet{He} proposed an initialization with a zero-mean Gaussian with variance equal to $2 / n_l$, taking into account the characteristics of the ReLU activation function.\\
In this work, we establish a connection between the weight initialization and the initial singular values of the weight matrices. These values have the potential to influence the final singular values and, consequently, oversmoothing. For this, we will use the circular law conjecture, which was proven with strong convergence by \citet{tao}.
\begin{theorem}[Circular Law Conjecture \citep{tao}]\label{theo:circular}
    Let $Z_n$ be a random matrix of order n, whose entries are i.i.d. samples of a random variable of zero-mean and bounded variance $\sigma_{std}^2$. Also let $\lambda_1,...,\lambda_n$ be the eigenvalues of $\frac{1}{\sigma_{std} \sqrt{n}} Z_n$. The circular law states that the distribution of $\lambda_i$ converges to a uniform distribution over the unit disk as n tends to infinity.
\end{theorem}
\noindent The circular law conjecture dictates the relationship between the standard deviation ($\sigma_{std}$) of the random variable and the radius of the disk. In fact, if we increase $\sigma_{std}$ there is a proportional increase of the disk radius, which in turn increases the largest eigenvalue of $Z_n$.

\section{Theoretical Analysis}
Despite their success in CNNs and FFNs, existing weight initialization methods fail to capture the effect of the graph topology, which is of high importance in GNNs. Therefore, we generalize the method developed in \citet{He} to provide new formulas about the variance flowing within the network, which takes into account the underlying graph topology. Subsequently, we simplify the formulas specifically for the case of a GCN and introduce a novel weight initialization method (G-Init) tailored for this particular model.

\subsection{Forward Propagation}
In order to simplify the notation, we will use the augmented normalized adjacency matrix (i.e., $\hat{A} = \hat{D}^{-1}(A + I)$) instead of the symmetrically normalized augmented adjacency matrix in Equation \ref{eq:general_gnn}. This simplification results in an equivalent analysis, differing only in the use of the factor $\frac{1}{d_i}$ instead of $\frac{1}{\sqrt{d_id_j}}$, in the formula of node representations. Hence, the representation of a node $i$ in layer $l$ becomes:
\begin{align*}\label{eq:gcn_aggr}
    y_{l}^{(i)} = \bigg(\frac{\alpha}{d_i}\sum\limits_{j \in \hat{N}(i)}{x_{l}^{(j)T}} + \beta \cdot x_{(0)}^{(i)} +
\end{align*}
\begin{equation}\label{eq:gcn_aggr}
    \gamma \cdot x_{(l-1)}^{(i)} \bigg) \cdot \left( \delta W_{l} + \epsilon I \right) + b_{l},
\end{equation}
where $N(i)$ is the neighborhood of node $i$ and $\hat{N}(i)$ is the augmented neighborhood, after self-loop addition, and $b_l$ is the bias. Here $x_{l}^{(j)}$ is an $n_l \times 1$ vector that contains the representation of node $j$ at layer $l$ and $W^{(l)}$ is an $n_l \times n_l$ matrix, where $n_l$ is the dimensionality of the layer $l$. Finally, $x_l^{(j)} = \sigma\left(y_{l-1}^{(j)}\right)$ , according to Equation \ref{eq:general_gnn}. Setting $x_l^{(i)'}=\frac{\alpha}{d_i}\sum\limits_{j \in \hat{N}(i)}{x_{l}^{(j)T}} + \beta \cdot x_{(0)}^{(i)} + \gamma \cdot x_{(l-1)}^{(i)}$ and $W_l' = \delta W_l + \epsilon I$ aligns Equation \ref{eq:gcn_aggr} above with Equation 5 in \citet{He}.\\
We let the initial elements of $W^{(l)}$ be drawn independently from the same distribution. Aligned with \citet{He}, we assume that the elements of $x_l^{(j)}$ are also mutually independent and drawn from the same distribution. Finally, we assume that $x_l^{(j)}$ and $W^{(l)}$ are independent of each other. Hence, we get:
\begin{equation}\label{eq:he_6}
    Var[y_l^{(i)}] = n_lVar[\delta w_lx_l^{(i)'} + \epsilon x_l^{(i)'}],
\end{equation}
where $y_l^{(i)}, x_l^{(i)'}$ and $w_l$ represent the random variables of each element in the respective matrices. We let $w_l$ have zero-mean, leading the variance to be:
\begin{align*}
    Var[y_l^{(i)}] = n_l \bigg( Var\left[\delta w_l x_l^{(i)'} \right] + Var\left[\epsilon x_l^{(i)'}\right] +\\
        2 Cov(\delta w_l x_l^{(i)'}, \epsilon x_l^{(i)'}) \bigg) = \\
        n_l\bigg( \delta^2 Var[w_l] E\left[\left(x_l^{(i)'}\right)^2\right] + \epsilon^2 Var\left[ x_l^{(i)'}\right] \bigg) \implies
\end{align*}
\begin{align}\label{eq:before-lemma}
    Var[y_l^{(i)}] \leq n_l \cdot E\left[\left(x_l^{(i)'}\right)^2\right] \left( \delta^2 Var[w_l] + \epsilon^2\right).
\end{align}
In Equation \ref{eq:before-lemma}, $x_l^{(i)'}$ aggregates information from the neighborhood of each node, its previous representation and its initial representation. These components are subsequently combined to generate the new node representation. Considering that $x_l^{(i)} = \sigma\left(y_{l-1}^{(i)}\right)$, it is necessary to decompose $x_l^{(i)'}$ into three components: the first containing $x_l^{(i)}$, the second containing $x_{l-1}^{(i)}$, and the third containing the remaining information of $x_l^{(i)'}$. In order to achieve that, we will employ the Cauchy–Bunyakovsky–Schwarz inequality (CBS), because it allows to transform a squared sum of elements into a sum of squares of these elements. 

\begin{lemma}\label{lemma:cauchy}
    \begin{align*}
    E\left[\left(x_l^{(i)'}\right)^2\right] = E\bigg[\bigg(\frac{\alpha}{d_i}\sum\limits_{j \in \hat{N}(i)}{x_{l}^{(j)T}} + \beta \cdot x_{(0)}^{(i)} + \\
    \gamma \cdot x_{(l-1)}^{(i)}\bigg)^2\bigg] \overset{CSB}{\leq} 
    \left(d_i + 1(\beta \neq 0) + 1(\gamma \neq 0)\right) \times \\
    \bigg(\frac{\alpha^2}{d_i^2} E\left[\left(x_l^{(i)}\right)^2\right] + 
    \gamma^2 \cdot E\left[\left(x_{l-1}^{(i)}\right)^2\right] + j(\alpha, \beta)\bigg),
    \end{align*}
    where $j(\alpha, \beta) = \frac{\alpha^2 \cdot k_l^{(i)}}{d_i^2} + \beta^2 \cdot E\left[\right(x_0^{(i)}\left)^2\right]$, $k^{(i)}_l = \sum\limits_{j \in N(i)}{\left(x_{l}^{(j)T}\right)^2}$, i.e., the sum of neighbor representations excluding the self representation and $1(\cdot)$ is the indicator function.
\end{lemma}
\noindent If we let $w_{l-1}$ have a symmetric distribution around zero and $b_{l-1} = 0$ then $y_{l-1}$ has zero-mean and symmetric distribution around its mean. This leads to $E\left[\left(x_l^{(i)}\right)^2\right] = \frac{1}{2}Var[y_{l-1}]$, when the activation function is  ReLU. Applying that result and Lemma \ref{lemma:cauchy} to Inequality \ref{eq:before-lemma} leads to the first main theorem of our work.

\begin{theorem}\label{theo:forward_theorem}
The upper bound of the variance of the signals flowing forward in a generic GNN defined by Equation \ref{eq:general_gnn} is:
\begin{align*}
    Var[y_l^{(i)}] \leq n_l \cdot \left(d_i + 1(\beta \neq 0) + 1(\gamma \neq 0)\right) \times \\
    \left(\frac{\alpha^2}{2d_i^2} Var[y_{l-1}^{(i)}] + \frac{\gamma^2}{2} \cdot Var[y_{l-2}^{(i)}] + j(\alpha, \beta)\right) \times
\end{align*}
\begin{align}\label{ineq:forwd_var}
\left(\delta^2 Var[w_l] + \epsilon^2 \right),
\end{align}

\noindent where $n_l$ is the weight matrix dimension, $d_i$ is the degree of node $i$, $1(\cdot)$ is the indicator function, $\alpha, \beta,\gamma,\delta$ and $\epsilon$ are parameters, depending on the underlying architecture of the model, and $j(\alpha, \beta)$ is defined in Lemma \ref{lemma:cauchy}.
\end{theorem}

\noindent The extended proof of Theorem \ref{theo:forward_theorem}, along with a lower bound of $Var[y_l^{(i)}]$, is available in Appendix A.

\subsection{Backward Propagation}
For back-propagation, the gradient for node $i$ is computed by:
\begin{equation}\label{eq:grad_eqn}
    \Delta x^{(i)}_l = W'_l \Delta y^{(i)}_l.
\end{equation}
We follow the same notation as in \citet{He}, where $\Delta x^{(i)}_l$ and $\Delta y^{(i)}_l$ denote gradients $\left(\frac{\partial{E}}{\partial{x^{(i)}_l}} \text{, } \frac{\partial{E}}{\partial{y^{(i)}_l}} \text{ and E is the loss}\right)$ and have dimensionality of $n_l$ $\times$ 1. Considering the common practice in GNNs to maintain a constant hidden dimension across layers, we proceed our analysis with $W_l$ being an $n_l$ $\times$ $n_l$ matrix. We note that, in the general case proposed in \citet{He}, $W'_l$ might be substituted by a matrix $\hat{W'}_l$, with $\hat{n}_l$ $\times$ $\hat{n}_l$ dimensions, which can be formed by $W'_l$ through reshaping. That modification does not affect our analysis, except in the appearance of the factor $\hat{n}_l$ in the results instead of $n_l$. More details about the use of $\hat{W'}_l$ can be found in \citet{He}.\\
We also set $\Delta x^{(i)'}_l = \frac{\alpha}{d_i} \sum\limits_{j \in \hat{N}(i)}{\left(\Delta x^{(j)T}_{l}\right)} + \gamma \cdot \Delta x_{(l-1)}^{(i)T}$, the average gradient reaching node $i$ derived from the forward pass and the interaction with its neighbors (message passing), plus the gradient of its representation in the previous layer. The presence of residual connections does not impact the gradient, as it adds a constant value to node representation (i.e., $x^{(i)}_0$).\\
If we assume that $w_l$ (a random variable representing the weight matrix elements) and $\Delta y^{(i)}_l$ are independent of each other and $w_l$ is initialized with a symmetric distribution around zero, then $\Delta x^{(i)}_l$ has zero mean for all $l$. Therefore, $\Delta x^{(i)'}_l$ also has a zero mean, as it results from the summation of zero-mean variables.\\
In back-propagation we have $\Delta y^{(i)}_l = \sigma'(y^{(i)}_l) \Delta x^{(i)'}_{l+1}$, where $\sigma'(\cdot)$ is the derivative of the activation function (i.e., ReLU), which is either one or zero, with equal probabilities. Additionally, we assume independence between $\sigma'(y^{(i)}_l)$ and $\Delta x^{(i)'}_{l+1}$. Hence, we arrive at the conclusion that $E\left[\Delta y^{(i)}_l\right] = E\left[\Delta x^{(i)'}_{l+1}\right] / 2 = 0$, due to the two branches of the ReLU derivative and the independence between $\sigma'(y^{(i)}_l)$ and $\Delta x^{(i)'}_{l+1}$. Consequently, $E\left[\left(\Delta y^{(i)}_l\right)^2\right] = Var\left[\Delta y^{(i)}_l\right] = \frac{1}{2} Var\left[\Delta x^{(i)'}_{l+1}\right]$. Finally, we compute the variance of the gradient in Equation \ref{eq:grad_eqn} as follows:
\begin{equation*}
    Var\left[\Delta x_l\right] = n_l Var[w'_l] Var\left[\Delta y^{(i)}_l\right]
\end{equation*}
\begin{equation}\label{eq:back_variance}
    = \frac{1}{2} n_l \bigg( \delta^2 Var[w_l] + \epsilon^2 \bigg) Var\left[\Delta x^{(i)'}_{l+1}\right].
\end{equation}
Following a similar approach as in the forward pass and in Lemma \ref{lemma:cauchy} we get:

\begin{lemma}\label{lemma:back_lemma}
    \begin{equation*}
        E\left[\left(\Delta x^{(i)'}_{l+1}\right)^2\right] = E\bigg[\bigg(\frac{\alpha}{d_i}\sum\limits_{j \in \hat{N}(i)}{\Delta x_{l+1}^{(j)T}} + \gamma \cdot \Delta x_{(l)}^{(i)T}\bigg)^2\bigg]
    \end{equation*}
    \begin{equation*}
        \overset{CSB}{\leq} \left(d_i + 1(\gamma \neq 0)\right) \cdot \bigg( \frac{\alpha^2}{d^2_i} E\left[\left(\Delta x^{(i)}_{l+1}\right)^2 \right] +
    \end{equation*}
    \begin{equation*}
        \gamma^2 E\left[\left(\Delta x^{(i)}_{l}\right)^2 \right] + q(\alpha) \bigg), 
    \end{equation*}
    where $q(\alpha) = \frac{\alpha^2}{d^2_i} \cdot o^{(i)}_{l+1}$, and $o^{(i)}_{l+1} = \sum\limits_{j \in N(i)}{\left(\Delta x^{(j)T}_{l+1}\right)^2}$, i.e., the sum of gradients originating from the neighbors of the node, excluding self-originating gradient and $1(\cdot)$ is the indicator function.
\end{lemma}

\noindent Taking into consideration that $Var\left[\Delta x^{(i)'}_{l+1}\right] = E\left[\left(\Delta x^{(i)'}_{l+1}\right)^2\right]$ and applying Lemma \ref{lemma:back_lemma} to Equation \ref{eq:back_variance} gives rise to the second main theorem of this work.

\begin{theorem}\label{theo:backward_theorem}
The upper bound of the variance of the gradients flowing backward in a generic GNN defined by Equation \ref{eq:general_gnn} is:
\begin{align}\label{eq:back_res}
    Var\left[\Delta x^{(i)}_l\right] \leq \frac{m_w}{1 - \gamma^2 m_w} \cdot \bigg( \frac{\alpha^2}{d^2_i} Var\left[\Delta x^{(i)}_{l+1} \right] + q(\alpha) \bigg), 
\end{align}
with
\begin{align*}
    m_w = \frac{1}{2}n_l \left(d_i + 1(\gamma \neq 0)\right) \cdot \bigg( \delta^2 Var[w_l] + \epsilon^2 \bigg),
\end{align*}

\noindent where $n_l$ is the weight matrix dimension, $d_i$ is the degree of node $i$, $1(\cdot)$ is the indicator function $\alpha, \gamma,\delta$ and $\epsilon$ are parameters, depending on the underlying architecture of the model, and $q(\alpha)$ is defined in Lemma \ref{lemma:back_lemma}.

\end{theorem}

\noindent The extended proof of Theorem \ref{theo:backward_theorem}, along with a lower bound of $Var[\Delta x_l^{(i)}]$, are provided in Appendix B.

\subsection{G-Init}
Having analyzed the generic case of convolutional GNNs, we focus on the simpler yet prominent GCN model. The GCN is defined by Equation \ref{eq:gcn}, a special case of Equation \ref{eq:general_gnn} with $\alpha = \delta = 1, \beta = \gamma = \epsilon = 0$. Leveraging Theorem \ref{theo:forward_theorem} and Theorem \ref{theo:backward_theorem}, we aim to control the variance of the input signal at the final layer ($L$) of the model and the variance of the gradients flowing backward within it.\\
Regarding the variance at the final layer, we utilize Equation \ref{ineq:forwd_var} and telescopically replace the factor $Var\left[y^{(i)}_{l-1}\right]$, until we arrive at $Var\left[y_{1}^{(i)}\right]$, which is the variance of the first layer of the model.
\begin{align*}
    Var\left[y_L^{(i)}\right] \leq \frac{n_L}{d_i}Var[w_L] \left(\frac{1}{2}Var\left[y_{L-1}^{(i)}\right] + k^{(i)}_L \right) \implies 
\end{align*}
\begin{align*}
    Var\left[y_L^{(i)}\right] \leq Var\left[y_1^{(i)}\right] \left( \prod\limits_{l=2}^L{\frac{n_l}{2d_i}Var[w_l]} \right) + 
\end{align*}
\begin{align}\label{ineq:result}
    \sum\limits_{l=2}^L{\left( \prod\limits_{j=l+1}^L{\frac{n_j}{2d_i}Var[w_j]}\right) \frac{n_l}{d_i} k_l^{(i)} Var[w_l]}.
\end{align}
A proper initialization method should avoid reducing or magnifying the magnitudes of input signals exponentially. We aim to control the upper bound of the variance of the final layer of the model, which in turn requires tuning the products of Equation \ref{ineq:result}, in order to obtain a proper scalar, e.g. 1. A sufficient condition to achieve this is the following:
\begin{equation}
    \frac{n_l}{2d_i}Var[w_l] = 1, \hspace{.7cm} \forall l.
\end{equation}
This leads to a zero-mean Gaussian distribution of the weights, with standard deviation (std) equal to $\sqrt{2d_i/n_l}$.\\
Similar results can be obtained for the weight initialization, based on the gradients flowing backward in the network, if we use the respective equations.\\
Using the initialization generated by either of the two directions (forward or backward) is sufficient, as both methods prevent the exponential reduction or increase of magnitudes in both the input signals (flowing forward) and the gradients (flowing backward). Hence, for the new initialization method (G-Init) we choose a zero-mean Gaussian, whose standard deviation is $\sqrt{2d_i/n_l}$.

\section{Experiments}


\subsection{Experimental Setup}
\textbf{Datasets:} Aligned to most of the literature about node classification tasks, we focus on some well-known benchmarks in the GNN domain: \textit{Cora, CiteSeer, Pubmed}, adopting the data splits of \citet{Kipf_gcn}, where all the nodes except the ones used for training and validation are used for testing. Additionally, we use the \textit{Arxiv} dataset \citep{arxiv_dataset} from the OGB suite. Further, we experiment with the \textit{Photo, Computers, Physics} and \textit{CS} datasets following the splitting method presented in \citet{rest_datasets}. Finally, we test the method on graph classification tasks, utilizing various datasets from two collections: TUDataset \citep{TUDataset} and MoleculeNet \citep{MoleculeNet}.\\
\noindent \textbf{Methods:} The focus of our experiments is on the effect of different initialization methods. For this reason, we combine the basic GCN architecture \citep{Kipf_gcn} with different weight initialization methods. Specifically, we compare our method (G-Init) against Xavier initialization \citep{Xavier} and Kaiming initialization \citep{He}. For each of the two competing methods, we explore two variants; namely drawing samples from a uniform distribution with predefined limits and drawing samples from a zero-mean Gaussian distribution of predefined standard deviation. We use the notation $Uniform$ and $Normal$ to denote these two variants. We also compare against a recently proposed initialization method for GNNs, i.e., VIRGO initialization \cite{gnn_init_1}.\\
\noindent \textbf{Hyperparameters:} We performed a hyperparameter sweep to determine the optimal hyperparameter values, based on their performance on the validation set. For GCN, the number of hidden units of each layer is set to 128 across all datasets, except for the \textit{Arxiv} dataset, where it is set to 256. $L_2$ regularization is applied with a penalty of $5 \cdot$$10^{-4}$, and the learning rate is set to $10^{-3}$.\\
\noindent \textbf{Configuration:} Each experiment is run 10 times and we report the average performance and standard deviations over these runs. We train all models for 200 epochs (1200 for \textit{Arxiv} dataset) using Cross Entropy as the loss function.

\subsection{Experimental Results}
The choice of an appropriate value of $d_i$ in G-Init is important. Choosing $d_i = 1$ would neglect the structure of the graph, while opting for a larger value could result in increased magnitudes of weight elements, leading to training instability. Consequently, we needed to identify a proper value for $d_i$ that balances these two factors. The smallest possible degree for a node in a graph, with self loops included, is equal to 2 (indicating a single neighbor). Empirically, we have found that values of $d_i$ in the range $(1, 2]$ achieve high performance. In our experiments, we set $d_i = 2$, except for the \textit{Arxiv} dataset where $d_i=1.6$ further improves the performance. In summary, we propose initializing the weights of a GCN with a zero-mean Gaussian distribution, with a standard deviation (std) of $\sqrt{4/n_l}$ (in the \textit{Arxiv} dataset, the std is equal to $\sqrt{3.2/n_l}$).\\ 

\begin{figure}[htp]
    \centering
    \includegraphics[width=.365\textwidth]{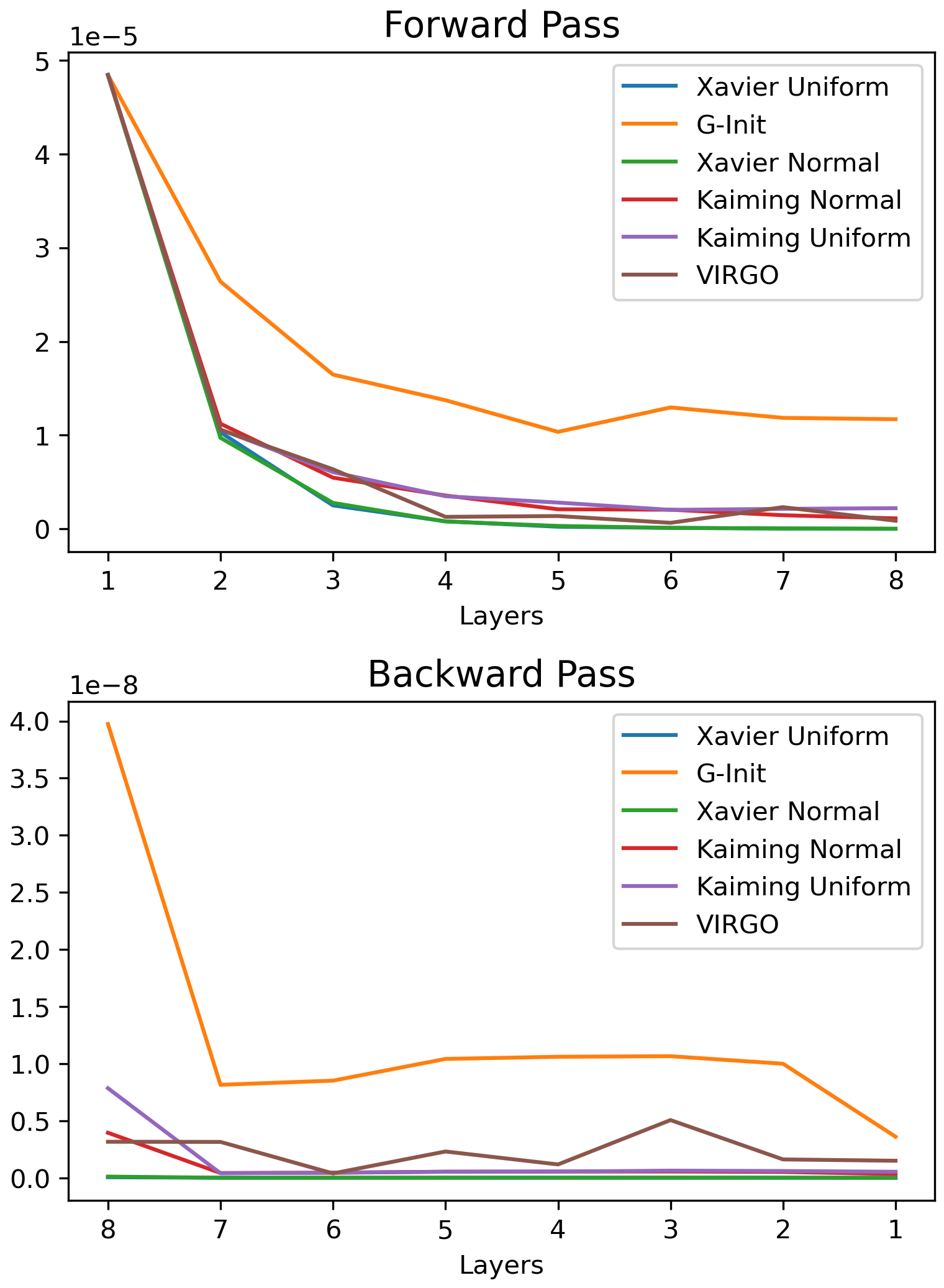}
    \caption{Variance plots on the \textit{Cora} dataset.}
    \vspace{0.6cm}
    \label{fig:variance_plots}
\end{figure}

\noindent \textbf{Variance stability:}\\
The first goal of our experimentation was to validate our theoretical results regarding the effect of G-Init on the variance within the model, before training. Specifically, we measured the average node variance as the signal flows forward in an 8-layer GCN and the average variance of the gradients flowing backward. Figure \ref{fig:variance_plots} presents the results for the \textit{Cora} dataset. Both in the forward and in the backward pass, G-Init maintains a larger variance than the other methods. \\

\begin{figure*}[ht]
    \centering
    \includegraphics[width=\textwidth]{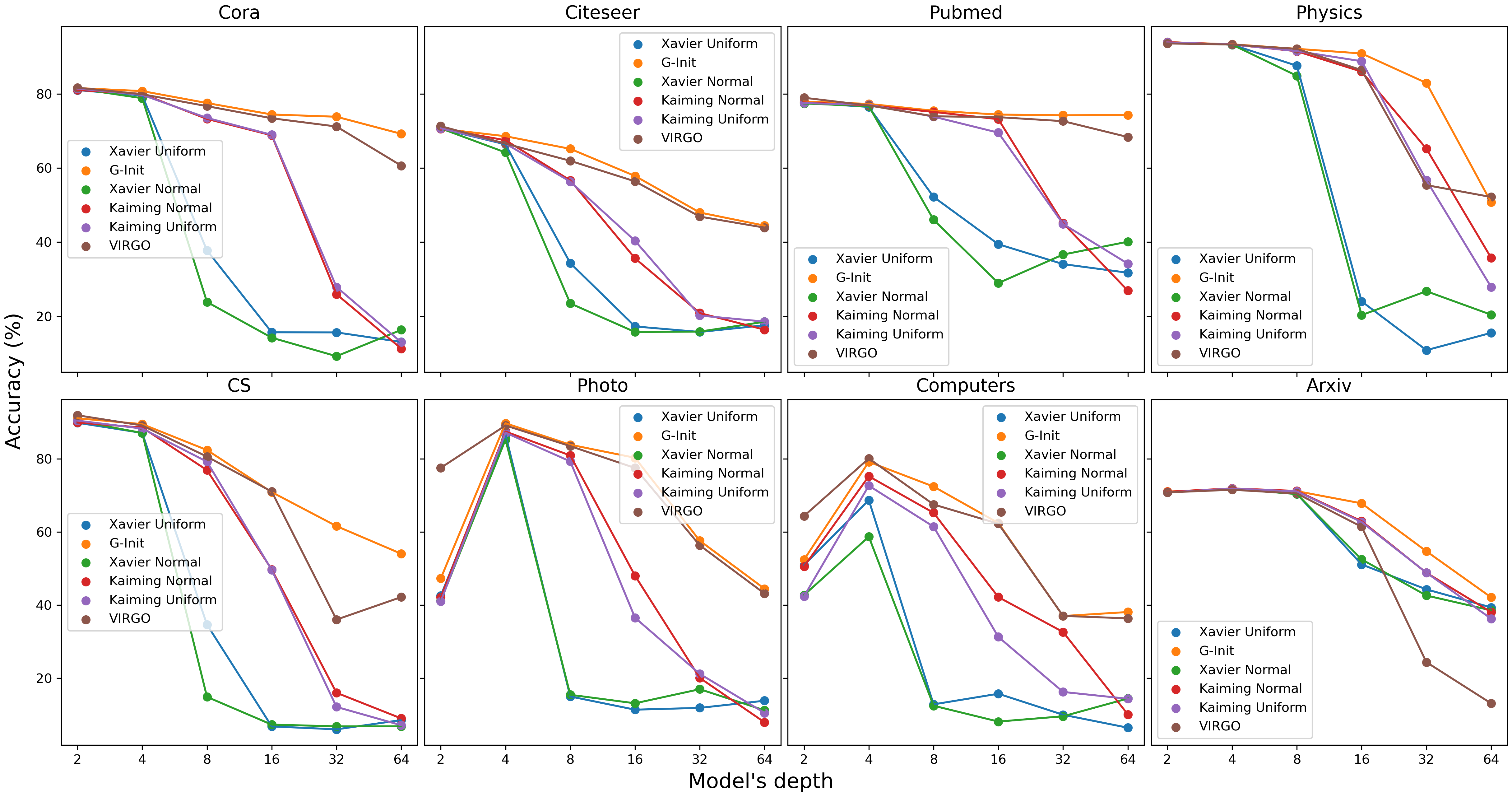}
    \caption{Comparison between 6 weight initialization methods across 8 datasets for varying GCN model depth.}
    \vspace{0.6cm}
    \label{fig:node_classification_all}
\end{figure*}

\noindent \textbf{Node classification and oversmoothing reduction:}\\
Given this increased variance in the signal and the gradients, the next goal of our experiments was to assess the effect on classification performance. Figure \ref{fig:node_classification_all} illustrates the performance of a GCN model with varying depth on 8 different datasets. G-Init enables the model to achieve superior performance across all combinations of datasets and depths. In fact, a GCN initialized with G-Init outperforms its counterparts initialized with any of the other methods. The experimental results verify the benefit of using a graph-informed weight intialization method, compared to the standard initialization methods that were devised for CNNs and FFNs.\\
GCN and models employing graph convolution, inherently reduce the variance of node representations (information signal), due to the repetitive application of the Laplacian operator. G-Init enables the model to maintain a higher variance in the lower layers, preventing the collapse of node representations to a subspace, where they would become indistinguishable.\\
Our experimentation also confirms the relationship between weight initialization and oversmoothing. GCN cannot avoid oversmoothing using classical initialization methods, even at moderate depths. On the contrary, G-Init significantly reduces this effect, facilitating the use of deeper architectures. This assertion is validated not only by the model's accuracy but also by the resulting t-SNE \citep{tsne} plots, which show that node representations have not been mixed to the extent of becoming indistinguishable. T-SNE plots for the \textit{Cora} dataset are presented in Figure \ref{fig:t_sne_Cora}, while for other datasets and initialization methods t-SNE plots are available in Appendix D.\\
The oversmoothing reduction achieved by G-Init can be attributed to different initial singular values of the weight matrices, compared to the classical initialization methods. In particular, \citet{He} propose an initialization using a zero-mean Gaussian with a standard deviation equal to $\sqrt{2}/\sqrt{n_l}$. Based on Theorem \ref{theo:circular}, we conclude that that approach creates weight matrices, whose eigenvalues lie on a disk with a radius equal to $\sqrt{2}$, while G-Init does the same on a disk of a greater radius (i.e., $\sqrt{4}$). Hence, G-Init initializes the model with weight matrices having larger maximum singular values than those produced by Kaiming initialization. These initial values may influence the largest singular values of the weight matrices after convergence, which, in turn, affect the extent to which oversmoothing can be reduced, based on Theorem \ref{theo:suzuki}.\\

\begin{figure}[ht]
    \centering
    \includegraphics[width=\columnwidth]{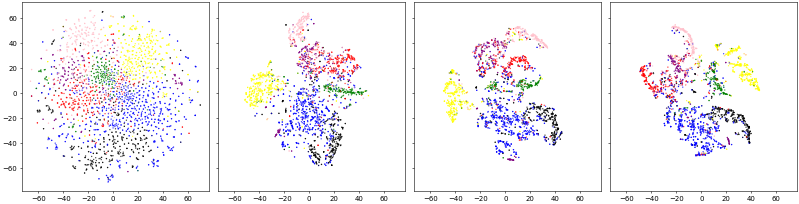}
    \includegraphics[width=\columnwidth]{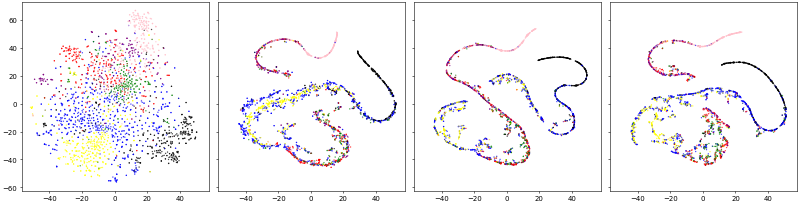}
    \caption{T-SNE plot of \textit{Cora} dataset.
    The upper row presents results for a G-Init initialized 32-layer GCN, while the lower row showcases results for a Kaiming Normal initialized 32-layer GCN.}
    \vspace{0.6cm}
    \label{fig:t_sne_Cora}
\end{figure}

\noindent \textbf{Performance under the ``cold start'' scenario:}\\
In order to further explore the impact of G-Init on oversmoothing, we conducted a set of experiments aiming to highlight the value of using deep architectures that are not prone to oversmoothing. Specifically, we introduced a ``cold start'' situation in the datasets (details about ``cold start'' are available in Appendix C), by replacing the feature vectors of the unlabeled nodes with all-zero vectors. For each dataset, Table \ref{tab:cold_start} presents the best performance achieved and the depth at which the model attains that performance, for G-Init, VIRGO, and Kaiming Normal initialization, because these three are the top-performing methods. An extended version of Table \ref{tab:cold_start} with results for every combination of initialization method and dataset is available in Appendix C. We observe that, G-Init consistently outperforms the other methods across almost all datasets.\\ 

\begin{table}[h]
    \centering
    \caption{Comparison of different initialization methods in the “cold start” scenario. Only the features of the nodes in the training set are available to the model. We present the best accuracy of the model and the depth (i.e. \# Layers) this accuracy is achieved.}
    \vspace{0.6cm}
    \resizebox{\columnwidth}{!}{
    \begin{tabular}{|c|p{2.7cm}|c c|}
        \hline
        Dataset & Method & Accuracy (\%) \& std & \#L\\
        \hline
        \multirow{3}{*}{Cora} & 
        Kaiming Normal & 68.35 \scriptsize$\pm$ 1.9 & 6 \\        
        {} & VIRGO & 73.01 \scriptsize$\pm$ 1.0 & 26 \\
        {} & G-Init & \textbf{74.04 \scriptsize$\pm$ 1.7} & 25 \\
        \hline
        \multirow{3}{*}{CiteSeer} &
        Kaiming Normal & 44.62 \scriptsize$\pm$ 1.8 & 6 \\        
        {} & VIRGO & 49.18 \scriptsize$\pm$ 1.4 & 18\\
        {} & G-Init & \textbf{49.75 \scriptsize$\pm$ 0.7} & 27 \\
        \hline
        \multirow{3}{*}{Pubmed} &
        Kaiming Normal & 68.48 \scriptsize$\pm$ 1.5 & 6 \\
        {} & VIRGO & 71.55 \scriptsize$\pm$ 1.5 & 14 \\
        {} & G-Init & \textbf{71.65 \scriptsize$\pm$ 1.8} & 23 \\
        \hline
        \multirow{3}{*}{Physics} &
        Kaiming Normal & \textbf{94.00 \scriptsize$\pm$ 0.0} & 2 \\
        {} & VIRGO & 82.34 \scriptsize$\pm$ 6.1 & 8\\
        {} & G-Init & 93.99 \scriptsize$\pm$ 0.0 & 1 \\
        \hline
        \multirow{3}{*}{CS} &
        Kaiming Normal & 90.13 \scriptsize$\pm$ 0.3 & 3 \\
        {} & VIRGO & 71.28 \scriptsize$\pm$ 1.9 & 6 \\
        {} & G-Init & \textbf{90.28 \scriptsize$\pm$ 0.2} & 3 \\
        \hline
        \multirow{3}{*}{Photo} &
        Kaiming Normal & 86.53 \scriptsize$\pm$ 0.6 & 5 \\
        {} & VIRGO & 83.00 \scriptsize$\pm$ 3.5 & 6\\
        {} & G-Init & \textbf{87.56 \scriptsize$\pm$ 1.2} & 4 \\
        \hline
        \multirow{3}{*}{Computers} &
        Kaiming Normal & 75.18 \scriptsize$\pm$ 3.0 & 4 \\
        {} & VIRGO & 75.17 \scriptsize$\pm$ 2.7 & 6 \\
        {} & G-Init & \textbf{78.03 \scriptsize$\pm$ 1.0} & 5 \\
        \hline
    \end{tabular}
    }
    \label{tab:cold_start}
\end{table}

\noindent \textbf{Graph classification:}\\
We also conducted experiments with various initialization methods on the task of graph classification, utilizing a GCN with average pooling on top. As shown in Table \ref{tab:graph_classification_all}, G-Init outperforms the other initialization methods in 8 out of 10 datasets, closely trailing as the second-best in the remaining two. These results verify that G-Init enhances the performance of GCN also in graph classification tasks.

\begin{table*}
    \centering
    \caption{Comparison between 6 weight initialization methods across 10 datasets for graph classification.}
    \vspace{0.6cm}
    \begin{tabular}{c c c c c c}
        \hline
        \multirow{2}{*}{Methods} & \multicolumn{5}{c}{Dataset} \\
        {} & BACE & BBBP & Clintox & Sider & Tox21 \\
        \hline
        \hline
         Xavier Normal & 81.07 \scriptsize$\pm$ 1.9 & 69.27 \scriptsize$\pm$ 1.4 & 91.51 \scriptsize$\pm$ 1.1 & 59.75 \scriptsize$\pm$ 1.2 & 75.33 \scriptsize$\pm$ 0.6 \\
         Xavier Uniform & 80.43 \scriptsize$\pm$ 2.5 & 69.68 \scriptsize$\pm$ 1.2 & 91.69 \scriptsize$\pm$ 1.1 & 59.77 \scriptsize$\pm$ 1.6 & 75.34 \scriptsize$\pm$ 0.8 \\ 
         Kaiming Normal & 80.78 \scriptsize$\pm$ 1.9 & 69.66 \scriptsize$\pm$ 1.2 & 91.59 \scriptsize$\pm$ 1.6 & 59.50 \scriptsize$\pm$ 1.5 & 75.52 \scriptsize$\pm$ 0.6 \\ 
         Kaiming Uniform & 79.36 \scriptsize$\pm$ 3.3 & 69.37 \scriptsize$\pm$ 1.0 & 91.15 \scriptsize$\pm$ 1.5 & 59.55 \scriptsize$\pm$ 1.8 & 75.47 \scriptsize$\pm$ 1.0 \\ 
         VIRGO & 81.16 \scriptsize$\pm$ 2.1 & 69.93 \scriptsize$\pm$ 1.2 & 92.26 \scriptsize$\pm$ 1.3 & 58.74 \scriptsize$\pm$ 1.0 & 74.97 \scriptsize$\pm$ 0.8 \\
         G-Init & \textbf{82.04 \scriptsize$\pm$ 1.1} & \textbf{70.03 \scriptsize$\pm$ 1.3} & \textbf{93.05 \scriptsize$\pm$ 1.4} & \textbf{60.01 \scriptsize$\pm$ 1.1} & \textbf{75.81 \scriptsize$\pm$ 1.0} \\ 
         \hline
    \end{tabular}
    
    \vspace{.2cm}    
    \begin{tabular}{c c c c c c}
        \hline
        \multirow{2}{*}{Methods} & \multicolumn{5}{c}{Dataset} \\
        {} & Toxcast & MUTAG & ENZYMES & Proteins & IMD$\text{B}_{\text{bin}}$\\
        \hline
        \hline
         Xavier Normal & 64.18 \scriptsize$\pm$ 0.5 & 90.53 \scriptsize$\pm$ 3.2 & 52.41 \scriptsize$\pm$ 2.6 & 68.84 \scriptsize$\pm$ 1.6 & 72.11 \scriptsize$\pm$ 1.0 \\ 
         Xavier Uniform & 64.20 \scriptsize$\pm$ 0.8 & 88.42 \scriptsize$\pm$ 3.2 & 51.25 \scriptsize$\pm$ 2.0 & 68.48 \scriptsize$\pm$ 1.3 & 72.67 \scriptsize$\pm$ 1.3\\ 
         Kaiming Normal & 64.45 \scriptsize$\pm$ 0.6 & 90.00 \scriptsize$\pm$ 1.8 & 52.41 \scriptsize$\pm$ 1.8 & 69.02 \scriptsize$\pm$ 1.5 & 72.33 \scriptsize$\pm$ 0.9\\ 
         Kaiming Uniform & 64.37 \scriptsize$\pm$ 0.3 & 89.47 \scriptsize$\pm$ 2.4 & \textbf{53.44 \scriptsize$\pm$ 2.6} & 68.93 \scriptsize$\pm$ 1.6 & 72.00 \scriptsize$\pm$ 1.5\\
         VIRGO & 64.40 \scriptsize$\pm$ 0.6 & 90.53 \scriptsize$\pm$ 3.2 & 49.67 \scriptsize$\pm$ 3.6 & \textbf{69.73 \scriptsize$\pm$ 1.7} & 72.00 \scriptsize$\pm$ 1.3 \\
         G-Init & \textbf{64.96 \scriptsize$\pm$ 0.5} & \textbf{92.63 \scriptsize$\pm$ 3.1} & 52.71 \scriptsize$\pm$ 2.5 & 68.84 \scriptsize$\pm$ 1.2& \textbf{72.89 \scriptsize$\pm$ 0.7} \\ 
         \hline
    \end{tabular}
    \label{tab:graph_classification_all}
\end{table*}

\begin{figure}[h]
    \centering
    \includegraphics[width=.99\columnwidth]{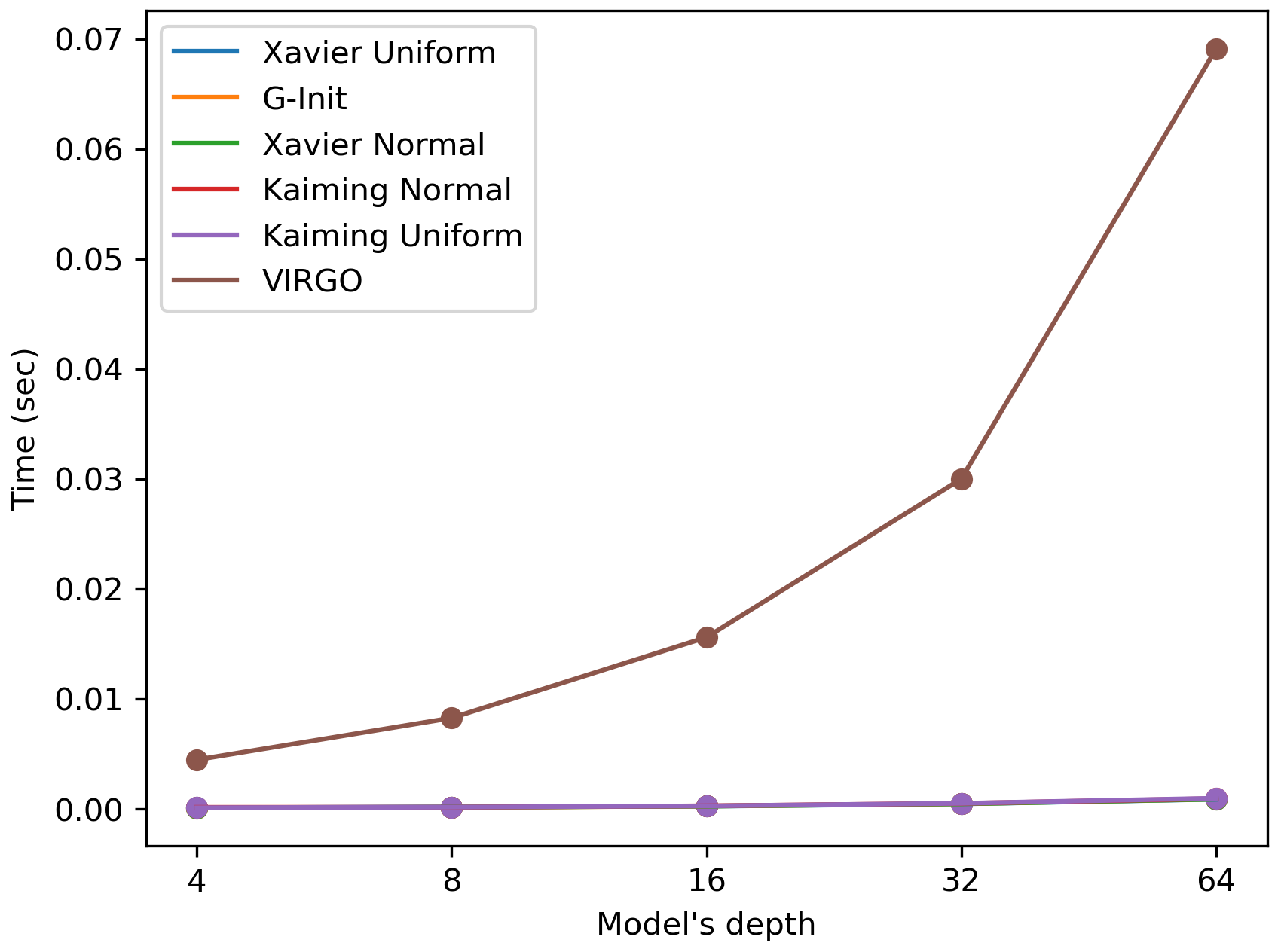}
    \caption{Comparison between 6 weight initialization methods for the needed time to initialize a GCN as its depth increases.}
    \vspace{0.6cm}
    \label{fig:virgo_time}
\end{figure}

~\\
\noindent \textbf{Comparison against VIRGO:}\\
The above results showed that VIRGO approached the performance of G-Init in some datasets. Since both have been specifically devised for GNNs, we compared them further. In particular, we examined the computational cost of the different methods, since VIRGO needs to calculate intermediate factors, which are depth dependent. Figure \ref{fig:virgo_time} presents the time needed to initialize a model as its depth increases. We observe that time required for initialization by VIRGO increases exponentially with the depth of the model, in contrast to G-Init. This behavior provides further reasons for using G-Init for GNN weight initialization.

\section{Related Work}
The most widely adopted weight initialization methods are those presented in \citet{Xavier} and \citet{He} and they have been extensively utilized across various types of models. \citet{Xavier} proposed their initialization method under the assumption of no activation function in the network, while \citet{He} focused on CNNs with ReLU activation functions.\\
More recent literature has examined initialization for GNNs. \citet{Jais} proposed an initialization method, based on a gradient flow analysis, tailored specifically to GCN, incorporating adaptive rewiring. An alternative was proposed by \citet{Han}, who utilized an MLP, trained only on the feature vectors of graph nodes, to initialize GNN weight matrices. Furthermore, \citet{gnn_init_1} proposed a weight initialization method, called VIRGO, which is based on the decomposition of node variance over message propagation paths and further decomposition of path variance. VIRGO has the overhead of calculating the product of the powers of the adjacency matrix up to the number of layers with an all-one vector, which results in repeated multiplications between a vector and a matrix. Additionally, in the context of graph classification tasks, VIRGO samples training graphs to approximate the underlying structure, which might not be representative, leading to a false perception of the structure and result in suboptimal initialization. Furthermore, we have empirically observed that VIRGO results are not stable across different seeds, when model depth increases. In contrast to these recent initialization methods, G-Init generalizes the ideas of Kaiming initialization to graph data, with minimal additional assumptions.\\ 
One of the important advantages of G-Init is its ability to reduce oversmoothing in deep GNNs. The problem of oversmoothing has attracted significant attention in recent literature. \citet{JK_Nets} introduced the concept of skip connections in GNNs through Jumping Knowledge Networks (JK-Networks), which combine information from lower layers with information reaching the uppermost layer. On the other hand, APPNP \citep{APPNP} and GCNII \citep{GCNII} introduced residual connections in GNNs, enabling the use of deep architectures. In a similar direction, DropEdge \citep{Dropedge} modified the network's topology to reduce oversmoothing. These methods are effective in reducing oversmoothing, but they are quite invasive, in comparison to G-Init, as they alter the structure of the GNN.\\
Part of the related work focuses on refining the weight initialization of GNNs, while another part is dedicated to oversmoothing reduction. Our analysis lies at the intersection of these research lines, presenting a novel weight initialization method, G-Init, and demonstrating its efficacy in oversmoothing reduction.

\section{Conclusion}
We conducted a theoretical analysis of the variance within a generic convolutional GNN and introduced a novel weight initialization method, extending the approach presented by \citet{He} to GNNs. In particular, the proposed method takes into account the graph topology and avoids exponentially large or small variance values. We assessed the behavior of the method through a series of experiments across a diverse set of benchmark datasets, encompassing both node and graph classification tasks. Our results have confirmed the relationship between weight initialization and oversmoothing reduction, which allowed us to use deep networks, without modifying either the model's architecture or the graph topology. Additional experiments under the ``cold start'' scenario, where features are limited to labeled nodes, demonstrated the superior performance of G-Init. Taking the proposed method further, we plan to explore layer-specific variations of G-Init and investigate how the depth of the model and other graph properties may influence the choice of the parameters of the method. 



\begin{ack}
The research work was supported by the Hellenic Foundation for
Research and Innovation (HFRI) under the 4th Call for HFRI PhD
Fellowships (Fellowship Number: 10860).
\end{ack}



\bibliography{mybibfile}



\onecolumn
\section*{Appendix}
\section*{Main Text Formulas}
The generic GNN equation as defined in main text is:
\begin{equation}\label{eq:general_gnn}
    H^{(l+1)} = \sigma\big(\big(\alpha \hat{A}H^{(l)} + \beta H^{(0)} + \gamma H^{(l-1)} \big) \times \big(\delta W^{(l)} + \epsilon I \big) \big),
\end{equation}
where $\alpha, \beta, \gamma, \delta \text{ and } \epsilon$ are preselected parameters that determine the convolutional GNN architecture.
\\

\noindent The variance of the gradient flowing backwards is defined as:
\begin{equation}\label{eq:back_variance}
    Var\left[\Delta x_l\right] = n_l Var[w'_l] Var\left[\Delta y^{(i)}_l\right]
    = \frac{1}{2} n_l \bigg( \delta^2 Var[w_l] + \epsilon^2 \bigg) Var\left[\Delta x^{(i)'}_{l+1}\right].
\end{equation}

\section*{Appendix A: Theorem 4 proof} \label{appdx:forward_theo_analysis}

\begin{theorem}
The upper bound of the variance of the signals flowing forward in a generic GNN defined by Equation \ref{eq:general_gnn} is:
\begin{align*}
    Var[y_l^{(i)}] \leq n_l \cdot \left(d_i + 1(\beta \neq 0) + 1(\gamma \neq 0)\right) \cdot \\
    \left(\frac{\alpha^2}{2d_i^2} Var[y_{l-1}^{(i)}] + \frac{\gamma^2}{2} \cdot Var[y_{l-2}^{(i)}] + j(\alpha, \beta)\right) \cdot
\end{align*}
\vspace*{-.5cm}
\begin{align}
\left(\delta^2 Var[w_l] + \epsilon^2 \right),
\end{align}

\noindent where $n_l$ is the weight matrix dimension, $d_i$ is the degree of node $i$, $1(\cdot)$ is the indicator function $\alpha, \beta,\gamma,\delta$ and $\epsilon$ are constants depending on the underlying architecture of the model, and $j(\alpha, \beta)$ is defined in Lemma 3.
\end{theorem}

\noindent \textbf{Proof:} 
We let the initialized elements in $W^{(l)}$ be mutually independent and have the same distribution. We also assume that elements in $x_l^{(i)}$ are also mutually independent and have the same distribution and finally we assume that $x_l^{(i)}$ and $W^{(l)}$ are independent of each other. Following a similar analysis as the one presented in \citet{He} we have:
\begin{equation}
    Var[y_l^{(i)}] = n_lVar[\delta w_lx_l^{(i)'} + \epsilon x_l^{(i)'}],
\end{equation}
where $y_l^{(i)}, x_l^{(i)'}$ and $w_l$ represent the random variables of each element in the respective matrices. We let $w_l$ have zero mean, hence the variance of the product is:
\begin{align}\label{eq:apdx_before-lemma}
    Var[y_l^{(i)}] = n_l \left( Var\left[\delta w_l x_l^{(i)'} \right] + Var\left[\epsilon x_l^{(i)'}\right] + 2 Cov(\delta w_l x_l^{(i)'}, \epsilon x_l^{(i)'}) \right)=  \\
    n_l\left( \delta^2 Var[w_l] E\left[\left(x_l^{(i)'}\right)^2\right] + \epsilon^2 Var\left[ x_l^{(i)'}\right] \right) \leq n_l \cdot E\left[\left(x_l^{(i)'}\right)^2\right] \left( \delta^2 Var[w_l] + \epsilon^2\right).
\end{align}
The special form of $x_l^{(i)'}$ is not directly related to $y_{l-1}^{(i)}$. In order to bridge that gap we will proceed our analysis with inequalities resulted by the usage of the Cauchy–Schwarz–Bunyakovsky inequality (CSB).
\begin{lemma}\label{lemma:apdx_cauchy}
    \begin{equation*}
        E\left[\left(x_l^{(i)'}\right)^2\right] = E\left[\left(\frac{\alpha}{d_i}\sum\limits_{j \in \hat{N}(i)}{x_{l}^{(j)T}} + \beta \cdot x_{(0)}^{(i)} + \gamma \cdot x_{(l-1)}^{(i)}\right)^2\right] \overset{CSB}{\leq} 
    \end{equation*}
    \begin{equation*}
        \left(d_i + 1(\beta \neq 0) + 1(\gamma \neq 0)\right) \left(\frac{\alpha^2}{d_i^2} E\left[\left(x_l^{(i)}\right)^2\right] + \frac{\alpha^2}{d_i^2} \sum\limits_{j \in N(i)}{\left(x_{l}^{(j)T}\right)^2} + \beta^2 \cdot E\left[\left(x_0^{(i)}\right)^2\right] + \gamma^2 \cdot E\left[\left(x_{l-1}^{(i)}\right)^2\right]\right)
    \end{equation*}
    \begin{equation*}
    = \left(d_i + 1(\beta \neq 0) + 1(\gamma \neq 0)\right) \left(\frac{\alpha^2}{d_i^2} E\left[\left(x_l^{(i)}\right)^2\right] + \gamma^2 \cdot E\left[\left(x_{l-1}^{(i)}\right)^2\right] + j(\alpha, \beta)\right),
    \end{equation*}
    where $k^{(i)}_l = \sum\limits_{j \in N(i)}{\left(x_{l}^{(j)T}\right)^2}$, i.e. sum of neighbors representations except of self representation and $j(\alpha, \beta) = \frac{\alpha^2 \cdot k_l^{(i)}}{d_i^2} + \beta^2 \cdot E\left[\right(x_0^{(i)}\left)^2\right]$.
\end{lemma}
\noindent Using Lemma \ref{lemma:apdx_cauchy}, Equation \ref{eq:apdx_before-lemma} transforms to:
\begin{equation}\label{ineq:apdx_after-CSB}
    Var[y_l^{(i)}] \leq n_l \cdot \left(d_i + 1(\beta \neq 0) + 1(\gamma \neq 0)\right) \left(\frac{\alpha^2}{d_i^2} E\left[\left(x_l^{(i)}\right)^2\right] + \gamma^2 \cdot E\left[\left(x_{l-1}^{(i)}\right)^2\right] + j(\alpha, \beta)\right) \cdot \left(\delta^2 Var[w_l] + \epsilon^2 \right).
\end{equation}
We have set $x_l^{(i)'}=\frac{\alpha}{d_i}\sum\limits_{j \in \hat{N}(i)}{x_{l}^{(j)T}} + \beta \cdot x_{(0)}^{(i)} + \gamma \cdot x_{(l-1)}^{(i)}$ and $W_l' = \delta W_l + \epsilon I$. Letting $w'_{l-1}$ (element of $W'_l$ matrix) have a symmetric distribution around zero and $b_{l-1} = 0$, then $y_{l-1}$ has zero mean and symmetric distribution around zero. In order to have symmetric distribution of $w'_{l-1}$ around zero we should have a symmetric distribution of $w_{l-1}$ around $-\epsilon$, based on the relationship between $W_l$ and $W'_l$.\\
This leads to $E\left[\left(x_l^{(i)}\right)^2\right] = \frac{1}{2}Var[y_{l-1}]$, when activation function is  ReLU. Applying that result to Inequality \ref{ineq:apdx_after-CSB} yields:
\begin{equation}
    Var[y_l^{(i)}] \leq n_l \cdot \left(d_i + 1(\beta \neq 0) + 1(\gamma \neq 0)\right) \left(\frac{\alpha^2}{2d_i^2} Var[y_{l-1}^{(i)}] + \frac{\gamma^2}{2} \cdot Var[y_{l-2}^{(i)}] + j(\alpha, \beta)\right) \cdot \left(\delta^2 Var[w_l] + \epsilon^2 \right).
\end{equation}
In a similar spirit, we also derive the lower bound for the variance. Considering that $x_l^{(i)} > 0$ for all $l,i$ we have the following lemma:

\begin{lemma}
    \begin{equation*}
        E\left[\left(x_l^{(i)'}\right)^2\right] = E\left[\left(\frac{\alpha}{d_i}\sum\limits_{j \in \hat{N}(i)}{x_{l}^{(j)T}} + \beta \cdot x_{(0)}^{(i)} + \gamma \cdot x_{(l-1)}^{(i)}\right)^2\right] \geq 
    \end{equation*}
    \begin{equation*}
        \frac{\alpha^2}{d_i^2} E\left[\left(x_l^{(i)}\right)^2\right] + \gamma^2 \cdot E\left[\left(x_{l-1}^{(i)}\right)^2\right] + j(\alpha, \beta).
    \end{equation*}
    This comes as a consequence of the fact that $\left(\sum\limits_j{m_j}\right)^2 \geq \sum\limits_j{m_j^2}$, for all $m_j \geq 0$.
\end{lemma}
\noindent Consequently, the lower bound of $Var\left[y_l^{(i)}\right]$ is given by:
\begin{equation}
    Var[y_l^{(i)}] \geq n_l \left(\frac{\alpha^2}{2d_i^2} Var[y_{l-1}^{(i)}] + \frac{\gamma^2}{2} \cdot Var[y_{l-2}^{(i)}] + j(\alpha, \beta)\right) \cdot \left(\delta^2 Var[w_l] + \epsilon^2 \right).
\end{equation}

\section*{Appendix B: Theorem 6 proof} \label{appdx:backward_theo_analysis}

\begin{lemma}\label{lemma:apdx_back_lemma}
    \begin{align*}
        Var\left[\Delta x^{(i)'}_{l+1}\right] = E\left[\left(\Delta x^{(i)'}_{l+1}\right)^2\right] = E\bigg[\left(\frac{\alpha}{d_i}\sum\limits_{j \in \hat{N}(i)}{\Delta x_{l+1}^{(j)T}} + \gamma \cdot \Delta x_{(l)}^{(i)T}\right)^2\bigg]
    \end{align*}
    \begin{align*}
        \overset{CSB}{\leq} \left(d_i + 1(\gamma \neq 0)\right) \cdot \bigg( \frac{\alpha^2}{d^2_i} E\left[\left(\Delta x^{(i)}_{l+1}\right)^2 \right] + \gamma^2 E\left[\left(\Delta x^{(i)}_{l}\right)^2 \right] + q(\alpha) \bigg), 
    \end{align*}
    where $q(\alpha) = \frac{\alpha^2}{d^2_i} \cdot o^{(i)}_{l+1}$, and $o^{(i)}_{l+1} = \sum\limits_{j \in N(i)}{\left(\Delta x^{(j)T}_{l+1}\right)^2}$, i.e., the sum of gradients originating from the neighbors of the node, excluding self-originating gradient and $1(\cdot)$ is the indicator function.
\end{lemma}

\begin{theorem}
The upper bound of the variance of the gradients flowing backward in a generic GNN defined by Equation \ref{eq:general_gnn} is:
\begin{align}
    Var\left[\Delta x^{(i)}_l\right] \leq \frac{m_w}{1 - \gamma^2 m_w} \cdot \bigg( \frac{\alpha^2}{d^2_i} Var\left[\Delta x^{(i)}_{l+1} \right] + q(\alpha) \bigg), 
\end{align}
with
\begin{align*}
    m_w = \frac{1}{2}n_l \left(d_i + 1(\gamma \neq 0)\right) \cdot \bigg( \delta^2 Var[w_l] + \epsilon^2 \bigg),
\end{align*}

\noindent where $n_l$ is the weight matrix dimension, $d_i$ is the degree of node $i$, $1(\cdot)$ is the indicator function $\alpha, \gamma,\delta$ and $\epsilon$ are constants depending on the underlying architecture of the model, and $q(\alpha)$ is defined in Lemma \ref{lemma:apdx_back_lemma}.

\end{theorem}

\noindent \textbf{Proof:} Using the fact that $E\left[\left(\Delta x^{(i)}_{l}\right)^2 \right] = Var\left[\Delta x^{(i)}_l\right]$ and Lemma \ref{lemma:apdx_back_lemma} to Equation \ref{eq:back_variance} we get:
\begin{align*}
    Var\left[\Delta x^{(i)}_l\right] = \frac{1}{2} n_l \bigg( \delta^2 Var[w_l] + \epsilon^2 \bigg) Var\left[\Delta x^{(i)'}_{l+1}\right] \leq \frac{1}{2} n_l \bigg( \delta^2 Var[w_l] + \epsilon^2 \bigg) \left(d_i + 1(\gamma \neq 0)\right) \cdot 
\end{align*}
\begin{align*}
    \bigg( \frac{\alpha^2}{d^2_i} Var\left[\Delta x^{(i)}_{l+1}\right] + \gamma^2 Var\left[\Delta x^{(i)}_l\right] + q(\alpha) \bigg)  = m_w \cdot \bigg( \frac{\alpha^2}{d^2_i} Var\left[\Delta x^{(i)}_{l+1}\right] + \gamma^2 Var\left[\Delta x^{(i)}_l\right] + q(\alpha) \bigg) \implies
\end{align*}
\begin{align*}
    Var\left[\Delta x^{(i)}_l\right] ( 1 - \gamma^2 m_w ) \leq m_w \cdot \bigg( \frac{\alpha^2}{d^2_i} Var\left[\Delta x^{(i)}_{l+1}\right] + q(\alpha) \bigg) \implies
\end{align*}
\begin{align*}
    Var\left[\Delta x^{(i)}_l\right] \leq \frac{m_w}{1 - \gamma^2 m_w} \cdot \bigg( \frac{\alpha^2}{d^2_i} Var\left[\Delta x^{(i)}_{l+1} \right] + q(\alpha) \bigg). 
\end{align*}

\noindent In the final inequality we have assumed that $m_w < \gamma^{-2}$. If this condition is not satisfied, then the last inequality has the inverse direction and establishes the lower bound for the variance of the gradients.\\

\noindent Following a similar approach as in Appendix \ref{appdx:forward_theo_analysis}, we derive the lower bound of the variance of the gradients.

\begin{lemma}
    \begin{align*}
        Var\left[\Delta x^{(i)'}_{l+1}\right] = E\left[\left(\Delta x^{(i)'}_{l+1}\right)^2\right] = E\bigg[\left(\frac{\alpha}{d_i}\sum\limits_{j \in \hat{N}(i)}{\Delta x_{l+1}^{(j)T}} + \gamma \cdot \Delta x_{(l)}^{(i)T}\right)^2\bigg] \geq
    \end{align*}
    \begin{align*}
        \frac{\alpha^2}{d^2_i} E\left[\left(\Delta x^{(i)}_{l+1}\right)^2 \right] + \gamma^2 E\left[\left(\Delta x^{(i)}_{l}\right)^2 \right] + q(\alpha).
    \end{align*}
    This comes as a consequence of the fact that $\left(\sum\limits_j{m_j}\right)^2 \geq \sum\limits_j{m_j^2}$, for all $m_j \geq 0$.
\end{lemma}

\noindent Consequently, the lower bound of $Var\left[\Delta x_l^{(i)}\right]$ is given by:
\begin{equation*}
    Var\left[\Delta x^{(i)}_l\right] \geq \frac{m_w}{\left(d_i + 1(\gamma \neq 0)\right)} \left( \frac{\alpha^2}{d^2_i} Var\left[\Delta x^{(i)}_{l+1}\right] + \gamma^2 Var\left[\Delta x^{(i)}_l\right] + q(\alpha) \right) \implies
\end{equation*}
\begin{align*}
    Var\left[\Delta x^{(i)}_l\right] \geq \frac{m'_w}{1 - \gamma^2 m'_w} \cdot \bigg( \frac{\alpha^2}{d^2_i} Var\left[\Delta x^{(i)}_{l+1} \right] + q(\alpha) \bigg) ,
\end{align*}
where $m'_w = m_w / \left(d_i + 1(\gamma \neq 0)\right)$ and once again we assume that $m'_w < \gamma^{-2}$. If this condition is not satisfied, then the last inequality has the inverse direction and establishes the upper bound for the variance of the gradients.

\section*{Appendix C: Extended analysis about the ``cold start" problem} \label{appdx:cold_start}

Oversmoothing primarily affects deep GNNs, raising the question of the necessity of deep architectures. Many benchmark datasets in the literature do not require deep networks due to the homophilic nature of the data. Homophily implies that the valuable information for the majority of the graph nodes is typically within close neighbors (2 or 3 hops away). One particular scenario where deeper architectures might be beneficial is the ``cold start" problem, where the majority of node features are missing, resembling to a recommender system encountering a new product or user. In this context, deeper GNNs may recover features from more distant nodes to generate informative representations. The ``cold start" datasets that we use in these experiments are generated by removing feature vectors from unlabeled nodes and replacing them with all-zero vectors.\\

\noindent We present an extended version of Table 1:

\begin{table}[H]
    \centering
    \caption{Comparison of different initialization methods on the “cold start” problem. We show accuracy percentage (\%) for the test set. Only the features of the nodes in the training set are available to the model. We also show at what depth (i.e. \# Layers) each model achieves its best accuracy.}
    \vspace{0.6cm}
    \begin{tabular}{|c|p{2.7cm}|c c|}
        \hline
        Dataset & Method & Accuracy (\%) \& std & \#L\\
        \hline
        \multirow{6}{*}{Cora} & Xavier Normal & 64.86 \scriptsize$\pm$ 0.7 & 4 \\
        {} & Xavier Uniform & 62.52 \scriptsize$\pm$ 3.9 & 4 \\
        {} & Kaiming Normal & 68.35 \scriptsize$\pm$ 1.9 & 6 \\
        {} & Kaiming Uniform & 62.78 \scriptsize$\pm$ 5.1 & 6 \\
        {} & VIRGO & 73.01 \scriptsize$\pm$ 1.0 & 26 \\
        {} & G-Init & \textbf{74.04 \scriptsize$\pm$ 1.7} & 25 \\
        \hline
        \multirow{6}{*}{CiteSeer} &
        Xavier Normal & 41.95 \scriptsize$\pm$ 0.2 & 4 \\
        {} & Xavier Uniform & 40.17 \scriptsize$\pm$ 5.0 & 5 \\
        {} & Kaiming Normal & 44.62 \scriptsize$\pm$ 1.8 & 6 \\
        {} & Kaiming Uniform & 44.20 \scriptsize$\pm$ 2.5 & 7 \\
        {} & VIRGO & 49.18 \scriptsize$\pm$ 1.4 & 18\\
        {} & G-Init & \textbf{49.75 \scriptsize$\pm$ 0.7} & 27 \\
        \hline
        \multirow{6}{*}{Pubmed} &
        Xavier Normal & 64.54 \scriptsize$\pm$ 0.9 & 4 \\
        {} & Xavier Uniform & 62.83 \scriptsize$\pm$ 2.8 & 4 \\
        {} & Kaiming Normal & 68.48 \scriptsize$\pm$ 1.5 & 6 \\
        {} & Kaiming Uniform & 65.93 \scriptsize$\pm$ 5.7 & 5 \\
        {} & VIRGO & 71.55 \scriptsize$\pm$ 1.5 & 14 \\
        {} & G-Init & \textbf{71.65 \scriptsize$\pm$ 1.8} & 23 \\
        \hline
        \multirow{6}{*}{Physics} &
        Xavier Normal & 94.00 \scriptsize$\pm$ 0.1 & 2 \\
        {} & Xavier Uniform & 93.98 \scriptsize$\pm$ 0.0 & 1 \\
        {} & Kaiming Normal & \textbf{94.00 \scriptsize$\pm$ 0.0} & 2 \\
        {} & Kaiming Uniform & 93.98 \scriptsize$\pm$ 0.0 & 1 \\
        {} & VIRGO & 82.34 \scriptsize$\pm$ 6.1 & 8\\
        {} & G-Init & 93.99 \scriptsize$\pm$ 0.0 & 1 \\
        \hline
        \multirow{6}{*}{CS} &
        Xavier Normal & 89.95 \scriptsize$\pm$ 0.2 & 1\\
        {} & Xavier Uniform & 90.17 \scriptsize$\pm$ 0.4 & 2 \\
        {} & Kaiming Normal & 90.13 \scriptsize$\pm$ 0.3 & 3 \\
        {} & Kaiming Uniform & 90.22 \scriptsize$\pm$ 0.4 & 2 \\
        {} & VIRGO & 71.28 \scriptsize$\pm$ 1.9 & 6 \\
        {} & G-Init & \textbf{90.28 \scriptsize$\pm$ 0.2} & 3 \\
        \hline
        \multirow{6}{*}{Photo} &
        Xavier Normal & 83.19 \scriptsize$\pm$ 5.1 & 5 \\
        {} & Xavier Uniform & 84.50 \scriptsize$\pm$ 1.4 & 3 \\
        {} & Kaiming Normal & 86.53 \scriptsize$\pm$ 0.6 & 5 \\
        {} & Kaiming Uniform & 87.11 \scriptsize$\pm$ 0.6 & 4 \\
        {} & VIRGO & 83.00 \scriptsize$\pm$ 3.5 & 6\\
        {} & G-Init & \textbf{87.56 \scriptsize$\pm$ 1.2} & 4 \\
        \hline
        \multirow{6}{*}{Computers} &
        Xavier Normal & 68.83 \scriptsize$\pm$ 6.5 & 4 \\
        {} & Xavier Uniform & 42.73 \scriptsize$\pm$ 6.7 & 2 \\
        {} & Kaiming Normal & 75.18 \scriptsize$\pm$ 3.0 & 4 \\
        {} & Kaiming Uniform & 72.42 \scriptsize$\pm$ 2.5 & 5 \\
        {} & VIRGO & 75.17 \scriptsize$\pm$ 2.7 & 6 \\
        {} & G-Init & \textbf{78.03 \scriptsize$\pm$ 1.0} & 5 \\
        \hline
    \end{tabular}
\end{table}

\section*{Appendix D: t-SNE plots} \label{appdx:tsne}

We present t-SNE \citep{tsne} plots for all datasets using a 32-layer GCN model initialized with the methods investigated in this study. T-SNE results are exhibited for layers 1, 9, 17 and 25. Specifically, we compare the t-SNE plots of a GCN initialized with the proposed G-Init method against the generally second-best performing initialization method, namely, Kaiming Normal. The t-SNE plots validate that our proposed method attains high accuracy by generating meaningful representations that reduce oversmoothing. In contrast, alternative initialization methods lead to the mixing of node embeddings, contributing to a degradation in performance. We observe this difference in the mixing of node representations as depth increases across the majority of datasets.

\begin{figure}[H]
    \centering
    \includegraphics[width=\textwidth]{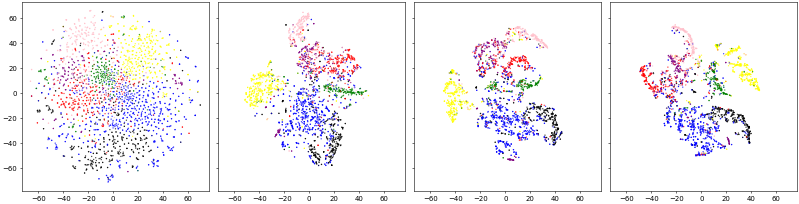}
    \includegraphics[width=\textwidth]{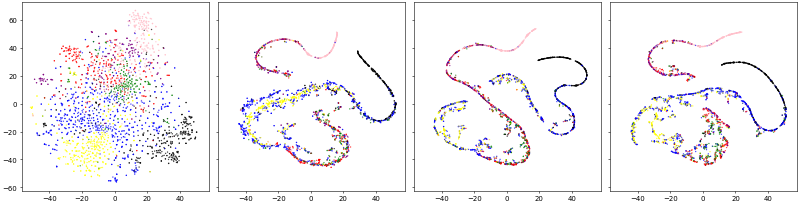}
    \caption{T-SNE plot of \textit{Cora} dataset.
    The upper row presents results for a G-Init initialized 32-layer GCN, while the lower row showcases results for a Kaiming Normal initialized 32-layer GCN.}
    \label{fig:t_sne_Cora}
    \end{figure}

    \begin{figure}[H]
    \centering
    \includegraphics[width=\textwidth]{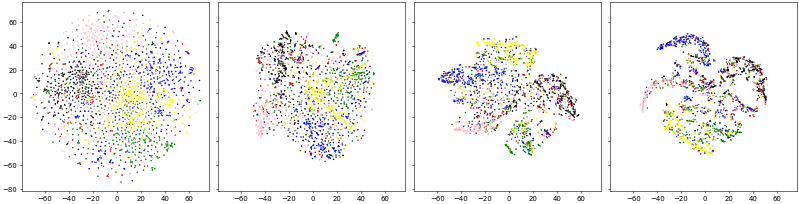}
    \includegraphics[width=\textwidth]{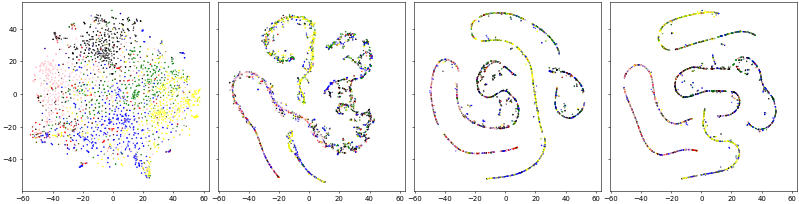}
    \caption{T-SNE plot of \textit{Citeseer} dataset.
    The upper row presents results for a G-Init initialized 32-layer GCN, while the lower row showcases results for a Kaiming Normal initialized 32-layer GCN.}
    \label{fig:t_sne_Citeseer}
    \end{figure}

    \begin{figure}[H]
    \centering
    \includegraphics[width=\textwidth]{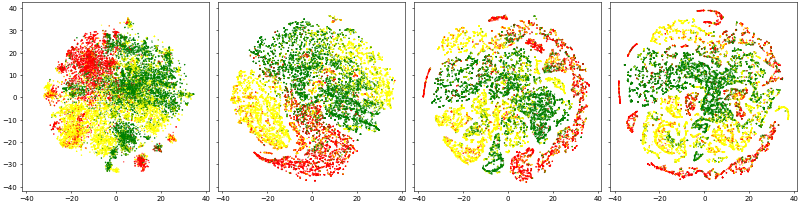}
    \includegraphics[width=\textwidth]{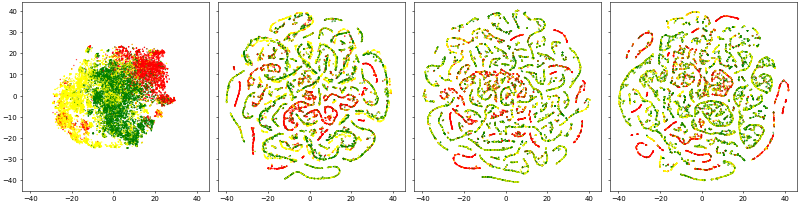}
    \caption{T-SNE plot of \textit{Pubmed} dataset.
    The upper row presents results for a G-Init initialized 32-layer GCN, while the lower row showcases results for a Kaiming Normal initialized 32-layer GCN.}
    \label{fig:t_sne_Pubmed}
    \end{figure}

    \begin{figure}[H]
    \centering
    \includegraphics[width=\textwidth]{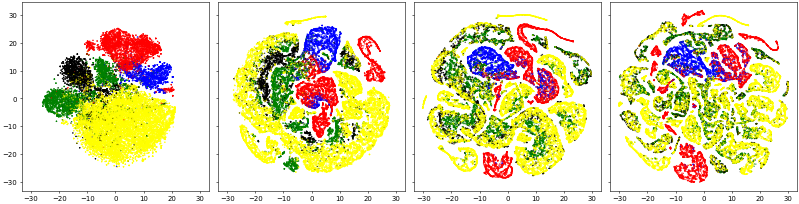}
    \includegraphics[width=\textwidth]{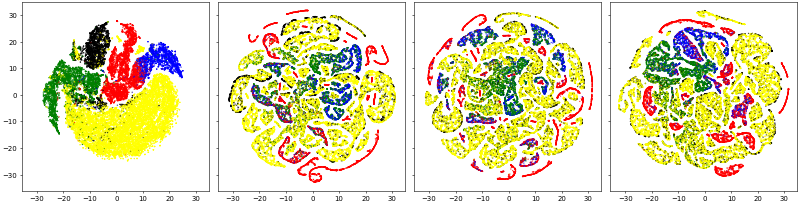}
    \caption{T-SNE plot of \textit{Physics} dataset.
    The upper row presents results for a G-Init initialized 32-layer GCN, while the lower row showcases results for a Kaiming Normal initialized 32-layer GCN.}
    \label{fig:t_sne_Physics}
    \end{figure}

    \begin{figure}[H]
    \centering
    \includegraphics[width=\textwidth]{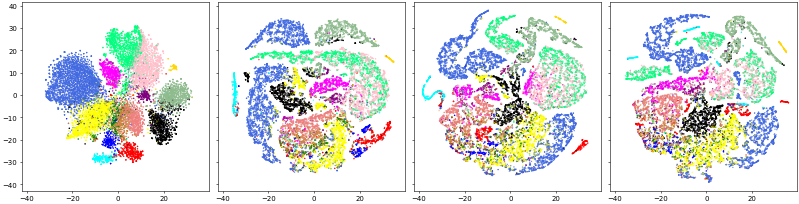}
    \includegraphics[width=\textwidth]{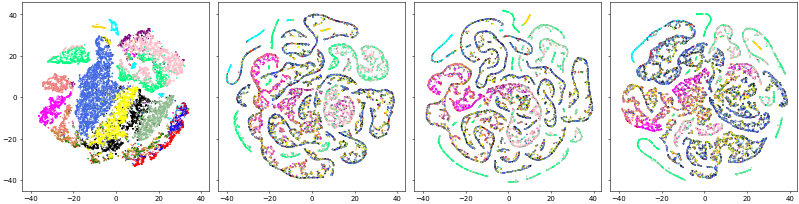}
    \caption{T-SNE plot of \textit{CS} dataset.
    The upper row presents results for a G-Init initialized 32-layer GCN, while the lower row showcases results for a Kaiming Normal initialized 32-layer GCN.}
    \label{fig:t_sne_CS}
    \end{figure}


\end{document}